\documentclass[sigconf]{acmart}
\AtBeginDocument{%
  }

\acmYear{2027}
\acmConference[WWW '27]{Proceedings of the ACM Web Conference 2027}{2027}{Dublin, Ireland}
\usepackage[ruled,vlined]{algorithm2e}
\usepackage{caption}

\usepackage{url}
\usepackage{algpseudocode}
\usepackage{amsmath}
\usepackage{multirow}
\usepackage{booktabs}
\usepackage{xcolor}
\usepackage[table]{xcolor}
\usepackage{newfloat}
\usepackage{balance}
\usepackage{listings}
\DeclareCaptionStyle{ruled}{labelfont=normalfont,labelsep=colon,strut=off} % DO NOT CHANGE THIS
\floatstyle{ruled}
\newfloat{listing}{tb}{lst}{}
\floatname{listing}{Listing}

\setcopyright{none}
\renewcommand\footnotetextcopyrightpermission[1]{}
\begin{document}

\title{MemCredit: From Relevance to Utility for Decision-Time Memory in Long-Horizon Web Agents}


\author{Quanjiang Guo}
\affiliation{%
  \institution{UESTC}
  \country{China}
}
\email{guochance1999@163.com}

\author{Sijie Wang}
\affiliation{%
  \institution{Nanyang Technological University}
  \country{Singapore}
}
\email{wang1679@e.ntu.edu.sg}

\author{Qianyang Zhu}
\affiliation{%
  \institution{UESTC}
  \country{China}
}
\email{zhuqianyang@std.uestc.edu.cn}

\author{Ben Zhang}
\affiliation{%
  \institution{UESTC}
  \country{China}
}
\email{zhangben@std.uestc.edu.cn}

\author{Jinchuan Zhang}
\affiliation{%
  \institution{UESTC}
  \country{China}
}
\email{jc.zhang@uestc.edu.cn}

\author{Ke Yan}
\affiliation{%
  \institution{UESTC}
  \country{China}
}
\email{kyan@uestc.edu.cn}

\author{zhao Kang}
\authornote{Corresponding author.}
\affiliation{%
  \institution{UESTC}
  \country{China}
}
\email{zkang@uestc.edu.cn}
% \renewcommand{\shortauthors}{Shuofu Liu, Quanjiang Guo, Xiao Liu, and Ying Liu}

\begin{abstract}
Large language model (LLM) agents increasingly operate over long Web interaction histories, making memory management critical for reliable and efficient decision making. Existing methods primarily optimize storage, retrieval, or compression, but the memories most semantically relevant to the current state are not necessarily those that most improve the agent's action. We introduce \textbf{MemCredit}, a decision-centric memory framework that shifts memory selection from \emph{semantic relevance} to \emph{decision utility}. MemCredit estimates the state-, reference-context-, and policy-conditional marginal contribution of individual memories through controlled counterfactual interventions, and distills these intervention-derived credits into a lightweight utility estimator. At inference time, semantic relevance is used only for high-recall candidate generation, while predicted utility arbitrates which memories are retained under a fixed token budget, without modifying the underlying agent policy or requiring additional environment interaction. We evaluate MemCredit on ALFWorld and WebShop against representative memory baselines. On ALFWorld, MemCredit achieves $86.57\%$ SR@50 under a 500-token historical-memory budget, compared with $65.67\%$ for BM25 retrieval. Counterfactual analysis further shows that memory utility is sparse, strongly state dependent, and poorly captured by semantic relevance. These results suggest that decision utility provides a more effective principle than relevance alone for long-horizon agent memory management. 
% Code is available at \url{https://anonymous.4open.science/r/memcredit/}.
\end{abstract}




\maketitle
\begin{figure}[t]
    \centering
    \includegraphics[width=\linewidth]{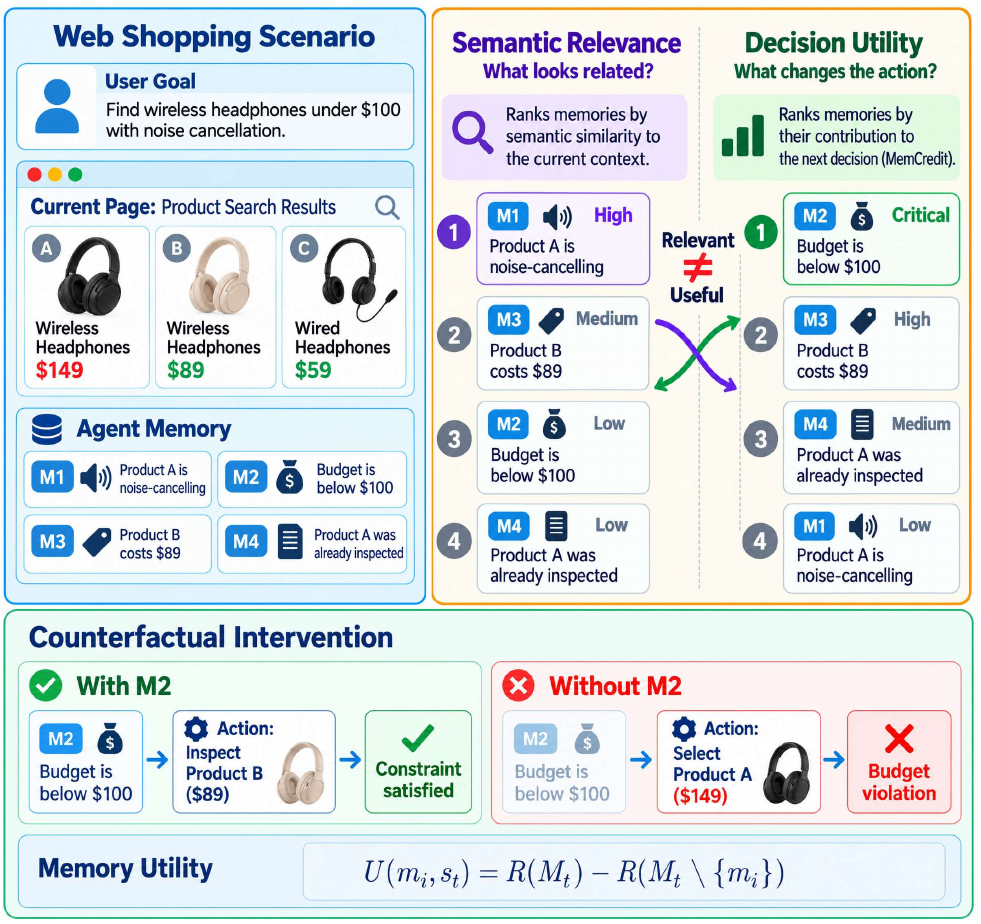}
    \caption{
    \textbf{Motivation of MemCredit.}
    Semantic relevance does not necessarily imply decision utility.
    A memory that appears less related to the current observation may be
    critical for avoiding an incorrect action, whereas a highly relevant
    memory can be redundant.
    }
    \label{fig:motivation}
\end{figure}

\section{Introduction}

Large language models (LLMs) are increasingly deployed as autonomous agents
that search, browse, invoke tools, and interact with online environments over
extended trajectories~\cite{yao2023react,shinn2023reflexion,zhao2024expel}.
Unlike single-turn generation, such interaction is inherently
\emph{stateful}: a correct decision may depend on an earlier user constraint,
previously observed information, or a failed action that should not be
repeated.
As trajectories grow, retaining the full interaction history becomes
increasingly inefficient, making memory management essential for long-horizon
agents~\cite{park2023generative,packer2023memgpt,zhong2024memorybank}.
Moreover, access to long context does not guarantee effective use of its
contents, as performance can vary substantially with how relevant information
is positioned and organized~\cite{liu2024lostmiddle}.

Existing memory systems address this challenge through recency-based
retention, semantic retrieval, trajectory compression, structured memory
banks, and learned memory policies~\cite{xu2025amem,zhu2026retrac,
chen2026lycheememory,luo2026multiturn}.
Despite these differences, memory selection is still commonly driven by
recency, semantic similarity, salience, or learned retrieval scores.
Such signals identify information that is \emph{related} to the current
state, but relatedness does not necessarily imply value for sequential
decision making.

As illustrated in Figure~\ref{fig:motivation}, a recently observed item may
be highly similar to the current observation yet contribute little to the
next action.
Conversely, an earlier constraint, failed attempt, or less similar piece of
evidence may be critical for avoiding an incorrect decision.
This reveals a fundamental mismatch between information retrieval and agent
memory management:
\textbf{semantic relevance measures whether a memory matches the current
state, whereas decision utility measures whether it improves the agent's
decision}.

We therefore ask:
\emph{would the agent make a worse decision if a particular memory were
unavailable?}
For a decision state $s_t$, reference memory context $S$, and candidate
memory $m_i$, we define its policy-conditional marginal decision utility as
\begin{equation}
\begin{split}
    U_\pi(m_i;s_t,S)
    &=
    \mathbb{E}
    \left[
        Q
        \mid
        s_t,S\cup\{m_i\},\pi_\theta
    \right]
    \\
    &\quad -
    \mathbb{E}
    \left[
        Q
        \mid
        s_t,S,\pi_\theta
    \right],
\end{split}
\label{eq:intro_utility}
\end{equation}

\begin{figure}[t]
    \centering
    \includegraphics[
        width=\columnwidth
    ]{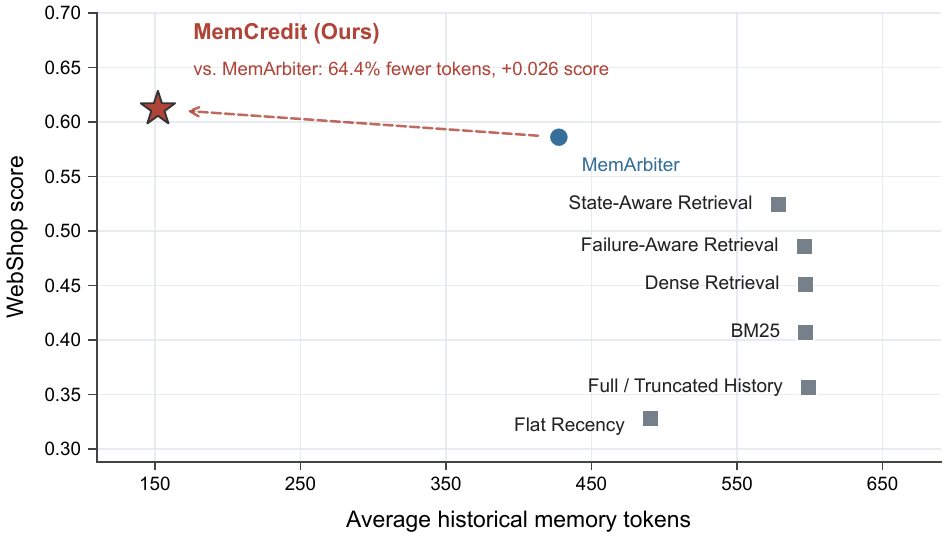}
    \caption{
    \textbf{Performance--memory Pareto comparison on WebShop.}
    MemCredit achieves the highest task score while using only $152.1$
    memory tokens per decision, about $2.8\times$ fewer than MemArbiter and
    approximately $3.9\times$ fewer than retrieval-based alternatives.
    The desirable region is the upper left: higher task performance with
    lower memory cost.
    }
    \label{fig:webshop_pareto}
\end{figure}

where $Q$ measures environment-grounded decision quality.
This formulation treats memory value as state-, reference-context-, and
policy-dependent, and casts memory management as a fine-grained
\emph{credit-assignment} problem rather than relevance ranking.
It is related to recent experience- and skill-based agents that evaluate
reusable knowledge through downstream outcomes~\cite{xia2026skillrl,
tu2026d2skill}, but operates at the granularity of individual memories.

Controlled intervention experiments support this distinction
(Sec.~\ref{sec:utility}; Table~\ref{tab:utility_summary} and
Figure~\ref{fig:utility_analysis}).
Across 1,041 context-conditional interventions from 45 ALFWorld tasks,
$86.9\%$
produce no measurable behavioral change, indicating that memory utility is
highly sparse.
Average utility rises to $0.176$ in failure contexts but is only $0.029$
outside failure states, while semantic relevance has almost no overall
correlation with counterfactual utility ($\rho=-0.011$).
These results suggest that semantic similarity is an unreliable proxy for
whether a memory is consequential to the current decision.

Motivated by these findings, we introduce \textbf{MemCredit}, a framework for
decision-time memory arbitration in long-horizon agents.
MemCredit decomposes interaction histories into atomic memories and uses
controlled counterfactual interventions to estimate their marginal decision
contributions.
The resulting credits supervise a lightweight utility estimator, amortizing
the cost of counterfactual evaluation.
At inference time, MemCredit separates \emph{candidate recall} from
\emph{memory arbitration}: semantic relevance retrieves a broad candidate
pool, while predicted utility determines which memories are retained under a
fixed token budget.
The underlying agent policy remains unchanged and requires no additional
environment interaction at deployment.

We evaluate MemCredit on ALFWorld~\cite{shridhar2021alfworld} and
WebShop~\cite{yao2022webshop}.
On ALFWorld, MemCredit achieves $86.57\%$ SR@50 under a 500-token
historical-memory budget, outperforming BM25 retrieval by $20.90$ percentage
points.
With a 750-token budget, it reaches $90.30\%$ SR@50, compared with $81.34\%$
for BM25, showing larger gains under tighter context constraints.

On WebShop, MemCredit achieves the highest task score of $0.612$ while using
only $152.1$ memory tokens per decision.
Compared with MemArbiter, the strongest non-MemCredit baseline, it improves
the score from $0.586$ to $0.612$ while using about $2.8\times$ fewer memory
tokens; relative to BM25, memory usage is reduced by approximately
$3.9\times$.
As shown in Figure~\ref{fig:webshop_pareto}, MemCredit yields a clear
performance--memory Pareto improvement and also reduces repeated searches,
invalid actions, and trajectory length.

Our contributions are threefold:
\begin{itemize}

    \item We formulate \textbf{decision-time memory credit assignment},
    distinguishing semantic relevance from the state- and policy-conditional
    marginal contribution of an individual memory, and show that such utility
    is sparse, state dependent, and poorly captured by relevance.

    \item We propose \textbf{MemCredit}, which estimates fine-grained memory
    credit through controlled counterfactual intervention and distills it
    into an amortized utility estimator for budget-aware memory arbitration.

    \item We demonstrate on ALFWorld and WebShop that separating
    relevance-based candidate recall from utility-based arbitration improves
    both task performance and context efficiency, especially under tight
    historical-memory budgets.
\end{itemize}

\begin{figure*}[ht]
	\centerline{\includegraphics[width=\linewidth]{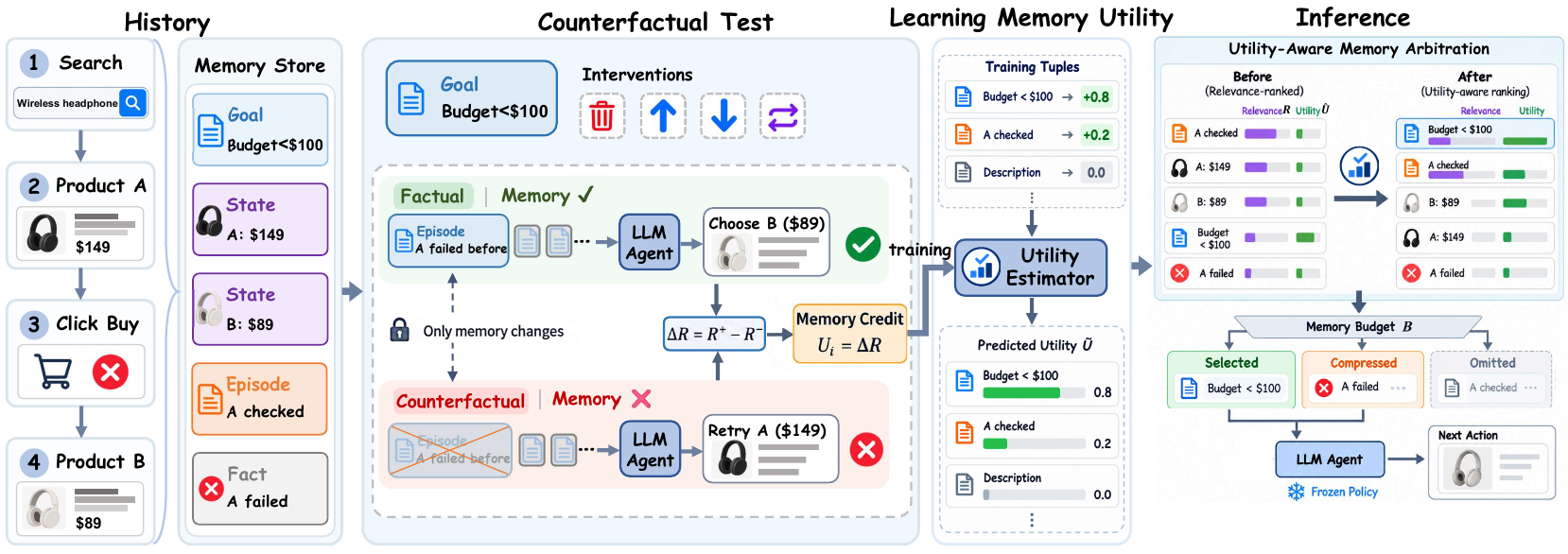}}
\caption{
\textbf{Overview of MemCredit.}
Interaction histories are organized into atomic memories, whose
policy-conditional decision utility is estimated through controlled
counterfactual interventions and distilled into a lightweight utility
estimator.
At inference time, semantic relevance provides high-recall candidates,
while predicted utility arbitrates memory selection under a fixed token
budget for the unchanged agent policy.
}
	\label{fig:methodology}
\end{figure*}

\section{Related Work}

\subsection{Agent Memory Management}

Agent memory supports long-horizon interaction through experience
retrieval~\cite{park2023generative}, long-term
storage~\cite{zhong2024memorybank}, tiered context
management~\cite{packer2023memgpt}, and evolving memory
structures~\cite{xu2025amem}.
Recent methods further compress trajectories into structured
states~\cite{zhu2026retrac}, learn bounded memory
updates~\cite{luo2026multiturn}, or select compressed blocks using
the query and working memory~\cite{chen2026lycheememory}.

MemArbiter~\cite{dong2026memarbiter} addresses the
\emph{Memory--Action Gap}: retrievable information may still fail
to guide actions.
It organizes history into functional banks and controls memory
presentation through demand, relevance, and temporal signals.
Its arbitration uses predefined signals rather than
intervention-derived memory credits.

Closer approaches explicitly evaluate memory utility.
CMI~\cite{srivastava2026cmi} scores memories through controlled
interventions in conversational tasks.
CICL~\cite{guan2026cicl} estimates decision-oriented utility and
assembles selected evidence into budgeted memory cards.
MeClear~\cite{yang2026meclear} combines leave-one-out screening
with sampled Shapley attribution to capture interactions and
suppress harmful memories.

MemCredit focuses on state-conditioned, item-level credit assignment
for sequential decisions.
It collects credits through paired replay from matched environment
states under a fixed action policy, then trains a lightweight estimator
for token-budgeted selection without additional runtime environment
interaction.
Single-removal attribution can mask redundancy among interacting
memories; Sec.~\ref{sec:arbitration} discusses how conditional reference
contexts mitigate this.

\subsection{Memory Utilization and Credit Assignment}

Effective memory use requires historical experience to improve downstream
behavior.
Reflexion~\cite{shinn2023reflexion} stores feedback-derived reflections in
episodic memory, ExpeL~\cite{zhao2024expel} extracts transferable insights
from agent trajectories, and SkillRL~\cite{xia2026skillrl} distills
trajectories into reusable skills.
More explicitly, D2Skill~\cite{tu2026d2skill} evaluates skills through paired
rollouts under the same policy, while memory-augmented RL optimizes memory
updates using trajectory-level rewards~\cite{luo2026multiturn}.
These approaches connect experience reuse with behavioral improvement, but
their credit signals mainly target reusable skills or memory-update
trajectories rather than individual memories at a specific decision state.

MemCredit instead performs \emph{decision-time credit assignment} for
individual memory items.
For memory $m_i$ at state $s_t$, it estimates marginal decision utility by
comparing decision quality with and without that memory under the same policy,
environment state, and reference context.
The resulting counterfactual credits supervise a lightweight utility
estimator for budget-aware arbitration, while leaving the agent policy
unchanged.
Semantic relevance is used for candidate recall, whereas predicted utility
determines final memory priority.

\section{Methodology}

MemCredit learns which historical memories improve a frozen agent's
current decision under a limited context budget.
As illustrated in Figure~\ref{fig:methodology}, the framework contains
four components: atomic memory construction, counterfactual credit
collection, utility estimation, and budget-aware arbitration.
The central idea is to estimate a memory's contribution
\emph{conditional on the other memories available at the decision}.
Offline interventions provide these credits; a lightweight estimator
then approximates them during context construction without further
environment interaction.

\subsection{Problem Formulation and Memory Representation}

At step $t$, an agent receives a task goal $g$, observation $o_t$,
and the previous interaction outcome $y_{t-1}$.
We denote this observable decision context by
$z_t=(g,o_t,y_{t-1})$, and distinguish it from the underlying
environment state $e_t$, which is used only for offline replay.
The history is represented by atomic memories
\begin{equation}
    M_t=\{m_i\}_{i=1}^{N_t},
    \qquad
    m_i=(c_i,b_i,\tau_i,\xi_i),
\end{equation}
where $c_i$ is content, $b_i$ is a functional type, $\tau_i$ is its
creation time, and $\xi_i$ records observable metadata such as source
actions, entities, and past outcomes.
Following MemArbiter~\cite{dong2026memarbiter}, the functional types
are \textsc{Goal}, \textsc{State}, \textsc{Constraint},
\textsc{Episodic}, and \textsc{Reference}.
These types organize the store without prescribing memory priority.

A selected set $S\subseteq M_t$ is serialized by a fixed renderer
$\mathcal{P}(S)$, and the frozen policy produces
\begin{equation}
    a_t\sim\pi_\theta\!\left(a\mid z_t,\mathcal{P}(S)\right).
\end{equation}
Let $L(S)$ be the token count of the complete rendered memory block,
including its labels and separators.
The historical-memory budget requires $L(S)\leq B$; the shared task
goal, current observation, and previous outcome are outside this budget.

Let $V_{\theta,H}(e_t,S)$ denote the expected environment return when
the current action is sampled with memory context $S$ and evaluated
over horizon $H$ under a fixed continuation protocol.
Memory selection seeks
\begin{equation}
    S_t^\star\in
    \arg\max_{S\subseteq C_t,\;L(S)\leq B}
    V_{\theta,H}(e_t,S),
    \label{eq:budget_selection}
\end{equation}
where $C_t$ is the retrieved candidate set.
This objective is not assumed to decompose into independent item values:
a memory may become redundant or useful depending on $S$.
The environment state and return are unavailable to the online selector;
they supply offline supervision only.

\subsection{Context-Conditional Counterfactual Credit Assignment}

\paragraph{\textbf{Conditional memory credit}.}
For candidate $m_i$ and reference context
$S\subseteq C_t\setminus\{m_i\}$, define
\begin{equation}
    \Delta_{i,t}(S)
    =
    V_{\theta,H}(e_t,S\cup\{m_i\})
    -
    V_{\theta,H}(e_t,S).
    \label{eq:memory_utility}
\end{equation}
Positive credit means that exposing $m_i$ improves the measured outcome
relative to the same reference context.
The credit is specific to the state, policy, reference context,
and evaluation horizon; it is not an intrinsic property of the memory.

Single-removal credit from the full candidate set is the special case
$S=C_t\setminus\{m_i\}$.
This case can mask redundancy: when two memories contain the same
necessary fact, removing either alone may have zero effect.
Conditioning on varied reference contexts exposes when an item adds
information and when that information is already represented.
Unlike Shapley attribution~\cite{ghorbani2019datashapley}, we predict
a conditional marginal effect rather than averaging over all coalition
orders to assign a global allocation of credit.
The intervention-based credit perspective is also related to
counterfactual credit assignment~\cite{foerster2018coma} and memory
attribution~\cite{yang2026meclear}.

\paragraph{\textbf{Budget-matched reference contexts}.}
We collect credits from contexts satisfying
$L(S\cup\{m_i\})\leq B$, matching the online selection constraint.
For a sampled target $m_i$, let $\ell_i=L(\{m_i\})$.
We draw an occupancy level
$\alpha\in\{0,\tfrac14,\tfrac12,\tfrac34,1\}$ and set
\begin{equation}
    b_i(\alpha)=\ell_i+\alpha(B-\ell_i).
\end{equation}
Starting with $m_i$, we randomly permute the remaining candidates and
insert each item only if the rendered context remains within
$b_i(\alpha)$.
If the resulting factual context is $A$, its reference context is
$S=A\setminus\{m_i\}$.
Items with $\ell_i>B$ are ineligible.
This sampling covers isolated memories, partial contexts, and
near-budget contexts without treating the full history as the only
reference.

\paragraph{\textbf{Paired environment replay}.}
For each $(e_t,z_t,S,m_i)$, both branches start from the same verified
environment snapshot and differ only in the rendered inclusion of $m_i$:
\begin{align}
    a_t^{+i}&\sim
      \pi_\theta(a\mid z_t,\mathcal{P}(S\cup\{m_i\})),\\
    a_t^{-i}&\sim
      \pi_\theta(a\mid z_t,\mathcal{P}(S)).
\end{align}
All other memory contents and their relative ordering are preserved.
The target is the effect of including a rendered item, which may
include length and position effects, rather than a content-only effect.

For a one-step evaluation, the observed credit is
\begin{equation}
    Y_{i,t}^{(1)}(S)
    =
    r_t^{+i}-r_t^{-i},
    \label{eq:one_step_utility}
\end{equation}
where $r$ is the specified environment reward.
For a longer horizon, each branch continues with the same fixed policy
and continuation protocol, giving
\begin{equation}
    Y_{i,t}^{(H)}(S)
    =
    \sum_{h=0}^{H-1}\gamma^h
    \left(r_{t+h}^{+i}-r_{t+h}^{-i}\right).
    \label{eq:rollout_utility}
\end{equation}
Rewards after termination are zero.
The intervention changes the current decision prompt; the continuation
protocol is shared across branches and fixed before collection.
Repeated paired runs, when used, are averaged to estimate
$\Delta_{i,t}(S)$; matched seeds control specified randomness but do
not imply deterministic GPU execution.
A factual rollout can be reused for multiple removals from the same $A$.

We sample states across ordinary progress, failures, repeated-action
risks, and subgoal transitions, and candidates across relevance levels
and functional types.
The resulting supervision is
\begin{equation}
    \mathcal{D}_{\mathrm{credit}}
    =
    \{(z_t,m_i,S,B,Y_{i,t}^{(H)}(S))\}.
\end{equation}
The collection protocol requires task-disjoint training, development,
and test sets.
Test-task outcomes are excluded from estimator fitting and threshold
selection.

\subsection{Learning Context-Conditional Memory Utility}

We amortize offline interventions with a lightweight estimator
\begin{equation}
    \widehat{\Delta}_{i,t}(S,B)
    =
    f_\phi\!\left(x(z_t,m_i,S,B)\right).
    \label{eq:utility_estimator}
\end{equation}
The features combine state--memory descriptors with descriptors of the
already available context:
\begin{equation}
    x=
    \left[
        x_{\mathrm{sm}}(z_t,m_i),
        x_{\mathrm{ctx}}(m_i,S,B)
    \right].
\end{equation}
The first group includes relevance, age, functional type, source,
entity overlap, observable progress, and history statistics.
The second includes maximum token-level Jaccard overlap with an item
in $S$, candidate-entity coverage by $S$, per-type context counts,
and budget occupancy $L(S)/B$.
Overlap and coverage are zero for the empty context.
These features expose whether a candidate contributes information
already present in the selected memories.

We retain an ExtraTrees regressor~\cite{geurts2006extratrees} and fit
the signed intervention targets with squared error:
\begin{equation}
    \mathcal{L}(\phi)
    =
    \frac{1}{|\mathcal{D}_{\mathrm{credit}}|}
    \sum_{d\in\mathcal{D}_{\mathrm{credit}}}
    \left(f_\phi(x_d)-Y_d\right)^2,
\end{equation}
where $x_d$ and $Y_d$ are the features and observed credit of sample $d$.
Positive, zero, and negative credits remain distinct training targets.
The predictor receives only information available before the decision:
reconstructed hidden states, future rewards, and intervention outcomes
are excluded from its inputs.
The estimator is frozen before evaluation.

At inference time, the same candidate can receive a different score as
$S$ changes.
The model therefore approximates the marginal value of adding a memory
to the current context rather than assigning it a fixed global rank.
Its fidelity is evaluated with task-grouped splits and relevance-only,
failure-only, context-blind, and shuffled-label controls.

\subsection{Utility-Aware Memory Arbitration}
\label{sec:arbitration}

\paragraph{\textbf{Candidate recall}.}
A lightweight relevance function retrieves a candidate pool
\begin{equation}
    q_{i,t}=\mathrm{Rel}(m_i,z_t),
    \qquad
    C_t=\mathrm{TopM}_{m_i\in M_t}\,q_{i,t}.
\end{equation}
Relevance limits the search space; it is not interpolated into the final
utility score.
Because omitted memories cannot be recovered during arbitration,
candidate-pool size is selected on development tasks and recall is
evaluated separately from final selection.

\paragraph{\textbf{Conditional budgeted selection}.}
Starting from $S_t^{(0)}=\varnothing$, MemCredit repeatedly predicts
candidate credits conditional on the current selected set.
At iteration $k$, the eligible candidates are
\begin{equation}
    \mathcal{F}_t^{(k)}
    =
    \left\{
    \begin{array}{l|l}
    m_i\in C_t\setminus S_t^{(k)}
    &
    \begin{array}{l}
    L(S_t^{(k)}\cup\{m_i\})\leq B,\\
    \widehat{\Delta}_{i,t}(S_t^{(k)},B)>\delta
    \end{array}
    \end{array}
    \right\},
\end{equation}
where the nonnegative threshold $\delta$ is selected using development
tasks.
For each eligible item, let
$d_i(S)=\max\{1,L(S\cup\{m_i\})-L(S)\}$.
The next item maximizes predicted credit per added token:
\begin{equation}
    \begin{aligned}
    i_k=
    \arg\max_{m_i\in\mathcal{F}_t^{(k)}}
    \frac{\widehat{\Delta}_{i,t}(S_t^{(k)},B)}
         {d_i(S_t^{(k)})},\\
    S_t^{(k+1)}=S_t^{(k)}\cup\{m_{i_k}\}.
    \end{aligned}
\end{equation}
Scores are recomputed after each insertion, allowing a candidate's
priority to decrease when related information has already been selected.
Selection stops when $\mathcal{F}_t^{(k)}$ is empty, so the budget is a
ceiling rather than a requirement to fill the context.

This procedure approximates Eq.~\ref{eq:budget_selection}; it does not
assume additive utilities or guarantee a globally optimal subset.
It can model local redundancy when covered by the training contexts,
but may miss complementary memories that are useful only jointly.
The final set $\widetilde{M}_t$ is serialized in a fixed chronological
order and passed to the unchanged action policy:
\begin{equation}
    a_t\sim
    \pi_\theta(a\mid z_t,\mathcal{P}(\widetilde{M}_t)).
\end{equation}
All online rescoring uses the frozen estimator, with at most
$O(|C_t|^2)$ candidate evaluations and no counterfactual rollouts or
additional environment calls.

\section{Experiments}

\subsection{Experimental Setup}

\paragraph{\textbf{Benchmarks}.}
We evaluate MemCredit on \textbf{ALFWorld} and \textbf{WebShop}.
For ALFWorld, following the evaluation protocol of MemArbiter~\cite{dong2026memarbiter}, we use all
134 tasks from the out-of-distribution evaluation split, with a maximum of
50 interaction steps per episode.
Its deterministic dynamics allow factual and counterfactual decisions to be
replayed from the same reconstructed environment state.
WebShop evaluates long-horizon Web interaction, where agents must preserve
user constraints, product attributes, and previous search outcomes.
We use a 50-task development set and report final evaluation on 200 tasks
over a 10K-product catalog.

\paragraph{\textbf{Models and Agent Protocol}.}
All main experiments use \textbf{Qwen3.6-27B-FP8} with deterministic decoding.
The primary Qwen policy is served with
vLLM~\cite{kwon2023pagedattention}.
At each step, every method receives the same task goal, current observation,
previous interaction outcome, and method-specific memory context.
The current observation is excluded from the memory budget, which constrains
historical information only.
For ALFWorld, admissible actions are provided to all methods, and we evaluate
historical-memory budgets of 500 and 750 tokens.
The agent policy, action parser, environment interface, and interaction limits
are kept fixed across methods.

For cross-model evaluation, we replace only the action-generation model with
\textbf{Qwen2.5-7B-Instruct}, \textbf{DeepSeek-V3}, or \textbf{GPT-5.4},
while retaining the same memory construction, state parsing, prompts, and
500-token memory budget.
Additional implementation and prompting details are provided in the Appendix.

\paragraph{\textbf{Compared Methods}.}
We compare MemCredit with
\textbf{Flat Recency}~\cite{dong2026memarbiter},
which retains recent history;
\textbf{BM25 Retrieval}~\cite{robertson2009probabilistic},
which ranks memories by lexical relevance;
\textbf{Dense Retrieval}~\cite{karpukhin2020dense},
which uses semantic similarity;
\textbf{Failure-Aware Retrieval},
a heuristic that prioritizes recent failure-related memories,
inspired by failure-aware experience
reuse~\cite{shinn2023reflexion,li2026fresh}; and
\textbf{State-Aware Retrieval},
a heuristic that selects memories aligned with the current
interaction state~\cite{wang2026samem}.
We also evaluate \textbf{MemArbiter}~\cite{dong2026memarbiter},
which controls memory presentation through functional banks,
dynamic signals, and a temporal gate.
Utility-only and relevance--utility hybrid variants of MemCredit
are evaluated in the ablation study.

\paragraph{\textbf{Counterfactual Credit Collection}.}
For ALFWorld, we collect \textbf{1,041 valid context-conditional
interventions} from \textbf{45 tasks}, following the protocol of the
Methodology section: reference contexts are sampled at occupancy levels
$\alpha\in\{0,\tfrac14,\tfrac12,\tfrac34,1\}$ under the 500-token budget,
and both branches are replayed from the same reconstructed environment
state with a fixed continuation protocol; every retained sample passes
the reconstruction check.
Supervision comes from task-disjoint training and development splits,
and test-task outcomes are excluded from estimator fitting and threshold
selection, as required by the collection protocol.
For WebShop, we apply the same protocol to a disjoint training set of
trajectories.
The resulting memory--state--utility triples train the utility estimator,
which is frozen before evaluation on the 200-task test set.
Further details are provided in the Appendix.

\paragraph{\textbf{Metrics and Evaluation}.}
For ALFWorld, we report \textbf{SR@50} overall and by task category, together
with post-failure recovery and failed-action repetition.
For WebShop, we report task performance, redundant interactions, and average
memory usage.
Counterfactual analysis measures utility sparsity, state dependence, and
relevance--utility correlation.
Utility prediction is evaluated with task-disjoint 5-fold GroupKFold using
Spearman correlation, AUPRC, and AUROC.
Confidence intervals are obtained by task- or episode-level bootstrap where
appropriate.
Additional details are provided in the Appendix.

\begin{table*}[ht]
\centering
\caption{
Performance on 134 unseen ALFWorld tasks under two historical-memory budgets.
We report SR@50 (\%) overall and across six task categories, with category sizes shown in parentheses.
Recovery and failed-action repetition characterize post-failure behavior.
}
\label{tab:alfworld_main}

\setlength{\tabcolsep}{3.5pt}
\renewcommand{\arraystretch}{1.12}

\begin{tabular}{c l cccccc c c c}
\toprule

\multirow{2}{*}{\shortstack[c]{\textbf{Memory}\\\textbf{Budget}}}
& \multirow{2}{*}{\textbf{Method}}
& \multicolumn{6}{c}{\textbf{Task-wise SR@50 (\%)}} 
& \multirow{2}{*}{\shortstack[c]{\textbf{Overall}$\uparrow$}}
& \multirow{2}{*}{\shortstack[c]{\textbf{Recovery}$\uparrow$}}
& \multirow{2}{*}{\shortstack[c]{\textbf{Failed Rep.}$\downarrow$}} \\

\cmidrule(lr){3-8}

& 
& \shortstack[c]{Pick\\(24)}
& \shortstack[c]{Look\\(18)}
& \shortstack[c]{Clean\\(31)}
& \shortstack[c]{Heat\\(23)}
& \shortstack[c]{Cool\\(21)}
& \shortstack[c]{Pick Two\\(17)}
& & & \\

\midrule

% ============================================================
% Budget = 500
% ============================================================

\multirow{9}{*}{500}

& Flat Recency
& 70.83 & 61.11 & 58.06 & 52.17 & 57.14 & 41.18
& 57.46 & 40.8 & 43.2 \\

& BM25
& 87.50 & 83.33 & 74.19 & 47.83 & 57.14 & 35.29
& 65.67 & 46.3 & 38.5 \\

& Dense Retrieval
& 87.50 & 83.33 & 77.42 & 60.87 & 66.67 & 47.06
& 71.64 & 53.7 & 32.1 \\

& Failure-Aware
& 83.33 & 88.89 & 74.19 & 56.52 & 66.67 & 47.06
& 70.15 & 59.6 & 27.6 \\

& State-Aware Retrieval
& 91.67 & 88.89 & 80.65 & 65.22 & 71.43 & 52.94
& 76.12 & 63.4 & 24.5 \\

& MemArbiter
& 95.83 & 94.44 & 83.87 & 73.91 & 80.95 & 64.71
& 82.84 & 68.2 & 22.4 \\

\cmidrule(lr){2-11}

& MemCredit (hybrid)
& 91.67 & 88.89 & 80.65 & 69.57 & 76.19 & 58.82
& 78.36 & 69.7 & 20.6 \\

& MemCredit (utility-only)
& 95.83 & 94.44 & 83.87 & 73.91 & 85.71 & 70.59
& 84.33 & 72.8 & 18.5 \\

\rowcolor{gray!15}
& \textbf{MemCredit}
& \textbf{95.83} & \textbf{100.00} & \textbf{87.10}
& \textbf{78.26} & \textbf{85.71} & \textbf{70.59}
& \textbf{86.57} & \textbf{75.9} & \textbf{16.4} \\

\midrule

% ============================================================
% Budget = 750
% ============================================================

\multirow{9}{*}{750}

& Flat Recency
& 79.17 & 72.22 & 64.52 & 60.87 & 66.67 & 47.06
& 65.67 & 44.5 & 38.0 \\

& BM25
& 95.83 & 94.44 & 83.87 & 65.22 & 90.48 & 52.94
& 81.34 & 55.8 & 31.0 \\

& Dense Retrieval
& 100.00 & 94.44 & 83.87 & 69.57 & 90.48 & 58.82
& 83.58 & 61.5 & 27.3 \\

& Failure-Aware
& 95.83 & 94.44 & 83.87 & 69.57 & 85.71 & 58.82
& 82.09 & 64.4 & 24.3 \\

& State-Aware Retrieval
& 100.00 & 94.44 & 87.10 & 73.91 & 90.48 & 58.82
& 85.07 & 67.1 & 22.1 \\

& MemArbiter
& 95.83 & 94.44 & 87.10 & 73.91 & 95.24 & 64.71
& 85.82 & 71.4 & 19.1 \\

\cmidrule(lr){2-11}

& MemCredit (hybrid)
& 100.00 & 100.00 & 87.10 & 73.91 & 95.24 & 70.59
& 88.06 & 72.8 & 17.9 \\

& MemCredit (utility-only)
& 100.00 & 94.44 & 90.32 & 78.26 & 95.24 & 70.59
& 88.81 & 74.7 & 16.5 \\

\rowcolor{gray!15}
& \textbf{MemCredit}
& \textbf{100.00} & \textbf{100.00} & \textbf{90.32}
& \textbf{82.61} & \textbf{95.24} & \textbf{70.59}
& \textbf{90.30} & \textbf{77.4} & \textbf{14.5} \\

\bottomrule
\end{tabular}

\end{table*}

\subsection{Overall Performance}
\label{sec:overall}

% WebShop main comparison (table*).

\begin{table*}[t]
\centering
\caption{
Overall performance on WebShop.
Score denotes the official WebShop task score, and memory tokens
are averaged over decision steps.
Lower memory usage, repeated-search rate, invalid-action rate,
and trajectory length are better.
}
\label{tab:webshop_main}

\setlength{\tabcolsep}{6pt}
\renewcommand{\arraystretch}{1.08}

\begin{tabular}{lccccc}
\toprule
\textbf{Method}
& \textbf{Score}$\uparrow$
& \textbf{Mem. Tokens} $\downarrow$
& \textbf{Repeat Search} $\downarrow$
& \textbf{Invalid Act.} $\downarrow$
& \textbf{Steps} $\downarrow$ \\
\midrule

\multicolumn{6}{l}{\textit{History-based}} \\
Full / Truncated History
    & 0.356
    & 599.1
    & 0.278
    & 0.353
    & 12.9 \\
Flat Recency
    & 0.328
    & 490.3
    & 0.292
    & 0.357
    & 13.0 \\

\midrule
\multicolumn{6}{l}{\textit{Relevance-based Retrieval}} \\
BM25
    & 0.407
    & 597.0
    & 0.260
    & 0.352
    & 12.8 \\
Dense Retrieval
    & 0.451
    & 596.8
    & 0.252
    & 0.346
    & 12.7 \\

\midrule
\multicolumn{6}{l}{\textit{State-aware / Structured Memory}} \\
Failure-Aware Retrieval
    & 0.486
    & 596.3
    & 0.257
    & 0.348
    & 12.7 \\
State-Aware Retrieval
    & 0.524
    & 578.4
    & 0.240
    & 0.335
    & 12.5 \\
MemArbiter
    & 0.586
    & 427.6
    & 0.227
    & 0.324
    & 12.3 \\

\midrule
\multicolumn{6}{l}{\textit{Decision-Utility-based}} \\
\rowcolor{gray!15} \textbf{MemCredit}
    & \textbf{0.612}
    & \textbf{152.1}
    & \textbf{0.213}
    & \textbf{0.316}
    & \textbf{12.0} \\

\bottomrule
\end{tabular}

\vspace{2pt}
\end{table*}

\paragraph{\textbf{ALFWorld}.}
Table~\ref{tab:alfworld_main} compares memory-management strategies under the
same Qwen3.6-27B-FP8 policy on 134 unseen ALFWorld tasks.
Under the tighter 500-token budget, MemCredit achieves \textbf{86.57\%}
SR@50, outperforming BM25, State-Aware Retrieval,
and MemArbiter.
This corresponds to a $20.90$-point gain over BM25.
The improvement spans all six task categories and is particularly pronounced
on \emph{Heat}, \emph{Cool}, and \emph{Pick Two}, where MemCredit improves
over BM25 by $30.43$, $28.57$, and $35.30$ points, respectively, indicating
the benefit of utility-aware arbitration for tasks with stronger historical
dependencies.

The advantage persists as the memory budget increases.
At 750 tokens, MemCredit reaches \textbf{90.30\%} SR@50, compared with
$81.34\%$ for BM25, $85.07\%$ for State-Aware Retrieval, and $85.82\%$ for
MemArbiter.
The margin over BM25 decreases from $20.90$ points at 500 tokens to $8.96$
points at 750 tokens, suggesting that explicit memory arbitration is
particularly valuable under tighter context constraints.
Behavioral metrics show the same pattern: at 500 tokens, MemCredit achieves
the highest recovery rate and the lowest failed-action repetition rate.
Overall, these results show that MemCredit improves both task success and
post-failure behavior through decision-aware memory selection.

\paragraph{\textbf{WebShop}.}
Table~\ref{tab:webshop_main} evaluates task performance and interaction
efficiency on 200 WebShop tasks over a 10K-product catalog.
MemCredit achieves the highest task score of \textbf{0.612}, outperforming
BM25, Dense Retrieval, State-Aware Retrieval,
and the strongest structured-memory baseline, MemArbiter.
At the same time, MemCredit uses only \textbf{152.1} historical memory tokens
per decision, compared with $427.6$ for MemArbiter and $597.0$ for BM25,
corresponding to reductions of \textbf{64.4\%} and \textbf{74.5\%},
respectively.

MemCredit also yields more efficient interaction behavior.
Compared with MemArbiter, repeated searches decrease from $0.227$ to $0.213$,
invalid actions from $0.324$ to $0.316$, and average trajectory length from
$12.3$ to $12.0$ steps.
Similar improvements are observed over relevance-based retrieval methods.
As visualized in Figure~\ref{fig:webshop_pareto}, MemCredit therefore shifts
the agent toward a more favorable performance--memory frontier, improving task
quality while substantially reducing the historical context required at each
decision step.

\begin{figure}[t]
    \centering
    \includegraphics[
        width=\linewidth
    ]{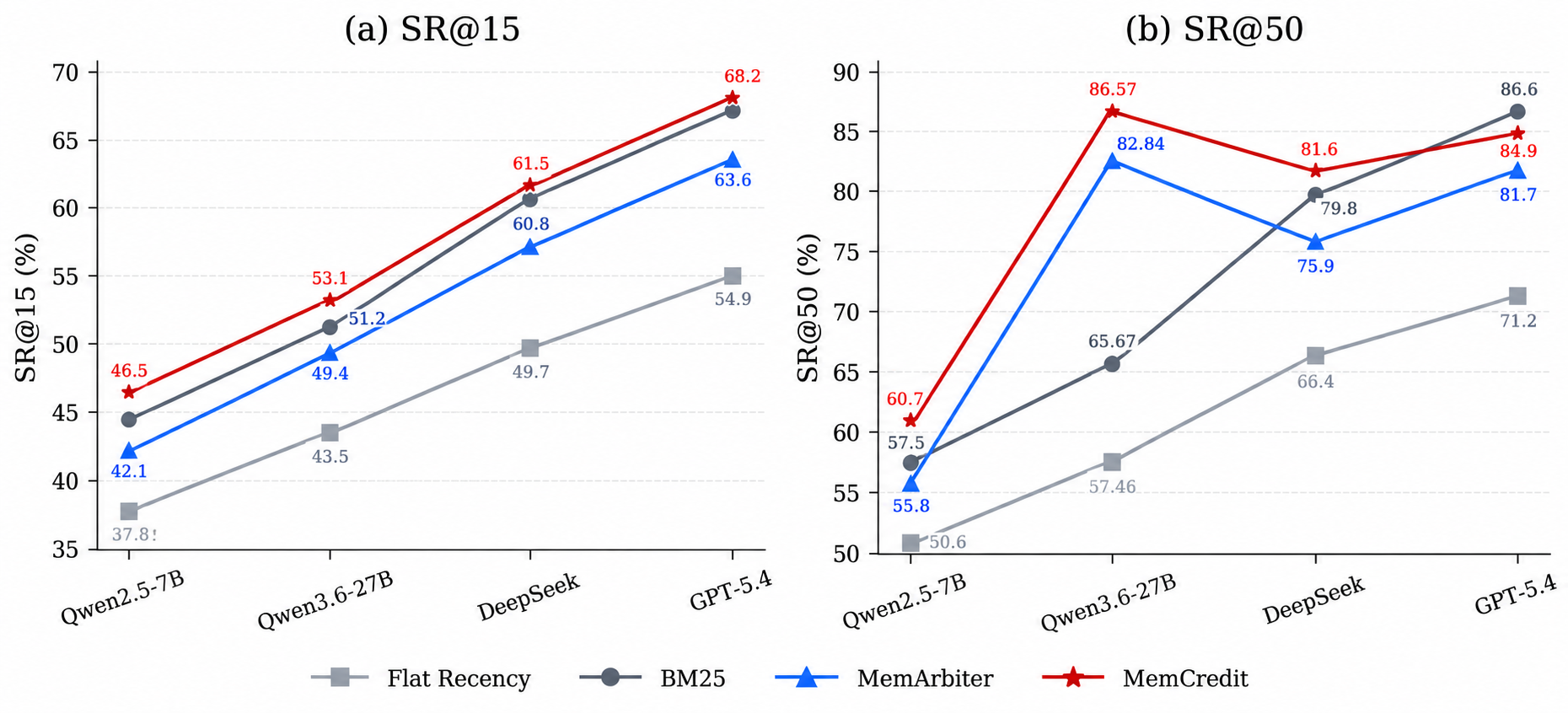}
    \caption{
    \textbf{Cross-model generalization on ALFWorld.}
    Cross-model comparison under a fixed 500-token memory budget for
    (a) SR@15 and (b) SR@50.
    Only the action-generation model is changed; all other agent components
    and evaluation settings remain fixed.
    }
    \label{fig:cross_model_radar}
\end{figure}

\paragraph{\textbf{Cross-model generalization}.}
We further examine whether MemCredit depends on the underlying
action-generation model.
We replace only the action model while keeping memory construction, state
parsing, prompts, interaction limits, and the 500-token memory budget fixed.
As shown in Figure~\ref{fig:cross_model_radar}, MemCredit outperforms
Flat Recency, BM25, and MemArbiter in almost all model and step-limit
settings across
\textbf{Qwen2.5-7B-Instruct}, \textbf{Qwen3.6-27B-FP8},
\textbf{DeepSeek-V3}, and \textbf{GPT-5.4}: the single exception is
SR@50 under GPT-5.4, where BM25 scores marginally higher.
Although absolute performance varies across backbones and interaction
horizons, the advantage of utility-aware memory arbitration holds across
nearly all settings, indicating that MemCredit is not tied to a specific
LLM.

\begin{table}[t]
\centering
\caption{
Summary of counterfactual memory-utility analysis.
Confidence intervals are obtained by task-clustered bootstrap.
}
\label{tab:utility_summary}
\small
\setlength{\tabcolsep}{4.0pt}
\renewcommand{\arraystretch}{1.05}

\begin{tabular}{lc}
\toprule
\textbf{Quantity} & \textbf{Estimate [95\% CI]} \\
\midrule

\multicolumn{2}{l}{\textit{Relevance--utility alignment}} \\

Combined Spearman $\rho$
& $-0.011~[-0.073,\,+0.077]$ \\

Dense-only Spearman $\rho$
& $-0.191$ \\

Broad-only Spearman $\rho$
& $+0.135~[+0.062,\,+0.209]$ \\

\midrule
\multicolumn{2}{l}{\textit{Utility concentration}} \\

Mean memory credit
& $+0.054~[+0.030,\,+0.080]$ \\

Zero-utility share
& $86.9\%$ \\

Failure-context utility
& $+0.176~[+0.077,\,+0.257]$ \\

Non-failure utility
& $+0.029~[+0.005,\,+0.057]$ \\

\bottomrule
\end{tabular}
\end{table}

\subsection{Understanding Decision Utility}
\label{sec:utility}

\begin{figure*}[t]
    \centering
    \includegraphics[width=0.9\linewidth]{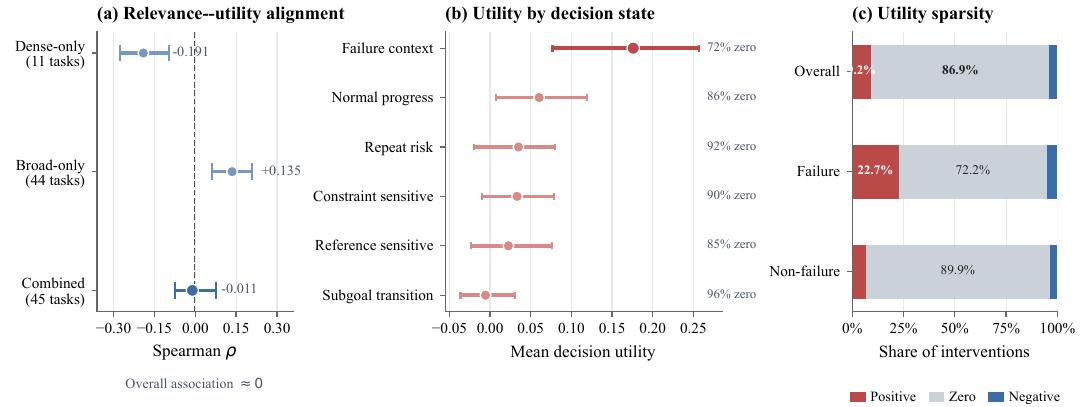}
    \caption{
    \textbf{Counterfactual analysis of memory decision utility.}
    (a) Relevance--utility association changes across task coverage and is
    near zero on the combined set.
    (b) Mean memory credit varies substantially across decision states,
    with the strongest effect after failures.
    (c) Decision utility is highly sparse: most interventions produce no
    measurable change.
    Error bars denote task-clustered 95\% bootstrap confidence intervals.
    }
    \label{fig:utility_analysis}
\end{figure*}

We next examine what counterfactual interventions reveal about the value of
individual memories at decision time.
Our causal set contains 1,041 valid context-conditional interventions from 45
ALFWorld tasks, all passing deterministic state reconstruction.
For each memory $m_i$ at state $s_t$ with reference context $S$, the
observed credit is
\[
U(m_i;s_t,S)
=
V(s_t,S\cup\{m_i\})-V(s_t,S),
\]
where positive credit indicates that exposing the memory improves the
decision; the removal-based statistics reported below are the
single-removal special case $S=C_t\setminus\{m_i\}$.
Table~\ref{tab:utility_summary} summarizes the main statistics, and
Figure~\ref{fig:utility_analysis} visualizes the corresponding trends.

\paragraph{\textbf{Relevance is not a stable proxy for decision utility}.}
Across the combined intervention set, semantic relevance has essentially no
overall association with marginal utility.
Moreover, the relationship changes with task coverage: the dense subset yields
$\rho=-0.191$, whereas the broader 44-task subset yields $\rho=+0.135$.
Thus, semantic similarity is not a stable estimate of a memory's marginal
decision value.

\paragraph{\textbf{Decision utility is sparse but consequential}.}
Mean credit is $0.054$, yet $86.9\%$ of
interventions produce no measurable change.
Most memories therefore have little effect at a given step, while a small
subset accounts for most of the observed impact.

\paragraph{\textbf{Utility is strongly state dependent}.}
Failure-context interventions have a mean utility of $0.176$, compared with $0.029$ outside failure states.
As shown in Figure~\ref{fig:utility_analysis}(b--c), failure contexts exhibit
the strongest effects, while several other state categories are substantially
weaker.

Overall, decision utility is \textbf{sparse, state dependent, and poorly
captured by semantic relevance}, motivating MemCredit to estimate memory
value explicitly.

\subsection{Can Decision Utility Be Predicted?}
\label{sec:prediction}

% Utility prediction (main text). NDCG omitted (sparse/tied labels make it
% unstable; details in the appendix).
\begin{table}[t]
\centering
\caption{
Task-disjoint prediction of intervention-derived decision utility using
5-fold GroupKFold by task ID.
The shuffled-label model serves as a negative control for spurious fitting.
}
\label{tab:utility_prediction}
\small
\setlength{\tabcolsep}{4.5pt}
\renewcommand{\arraystretch}{1.05}

\begin{tabular}{lccc}
\toprule
\textbf{Feature Set}
& \textbf{Spearman} $\uparrow$
& \textbf{AUPRC} $\uparrow$
& \textbf{AUROC} $\uparrow$ \\
\midrule

Relevance-only
& $-0.031$
& 0.097
& 0.483 \\

Content-only
& $+0.016$
& 0.122
& 0.493 \\

Failure-only
& $+0.095$
& 0.184
& 0.658 \\

\textbf{Full}
& $\mathbf{+0.220}$
& $\mathbf{0.270}$
& $\mathbf{0.744}$ \\

\midrule
Shuffled labels
& $+0.048$
& 0.146
& -- \\

\bottomrule
\end{tabular}
\end{table}

We next ask whether decision utility can be predicted from information
available at inference time.
Table~\ref{tab:utility_prediction} reports task-disjoint 5-fold GroupKFold
evaluation by task identity.

\paragraph{\textbf{Utility is predictable beyond relevance and failure cues}.}
Relevance-only features provide little predictive signal
($\rho=-0.031$, AUPRC $0.097$, AUROC $0.483$), while failure-related
features are more informative (AUPRC $0.184$, AUROC $0.658$).
The full estimator performs best, reaching $\rho=0.220$, AUPRC $0.270$,
and AUROC $0.744$.
This indicates that useful-memory prediction depends on signals beyond
semantic similarity or an explicit failure indicator.
A context-blind control that omits the context descriptors
$x_{\mathrm{ctx}}$ isolates the contribution of conditional context
features (Appendix).

\paragraph{\textbf{Prediction remains imperfect}.}
Although the full estimator consistently outperforms relevance-only,
failure-only, and shuffled-label controls, its moderate rank correlation
shows that fine-grained utility prediction remains challenging.
MemCredit therefore uses the estimator as a practical ranking signal rather
than an exact regressor of counterfactual value.

We omit NDCG@5 from the main analysis because zero-inflated utility labels
produce unstable rankings under ties; additional diagnostics are reported in
the Appendix.

\subsection{Ablation and Analysis}
\label{sec:ablation}

% Ablations (main text), ALFWorld budget 500 unless noted. v2 corrected numbers.
\begin{table}[t]
\centering
\caption{
Ablation of MemCredit on ALFWorld under a 500-token memory budget.
$\Delta$ is measured relative to utility-only ranking with full features.
}
\label{tab:ablation}
\small
\setlength{\tabcolsep}{4.2pt}
\renewcommand{\arraystretch}{1.05}

\begin{tabular}{lc}
\toprule
\textbf{Variant} & \textbf{SR@50 (\%)} $\uparrow$ \\
\midrule

\multicolumn{2}{l}{\textit{Scoring strategy}} \\

Relevance-only (BM25)
& 65.67 ($-18.66$) \\

Failure-only heuristic
& 70.15 ($-14.18$) \\

Relevance + utility ($\lambda=0.5$)
& 78.36 ($-5.97$) \\

Utility-only (full features)
& 84.33 (\phantom{$+$}0.00) \\

\midrule
\multicolumn{2}{l}{\textit{Utility features}} \\

\textbf{MemCredit (w/o explicit failure/state)}
& \textbf{86.57} ($+2.24$) \\

\bottomrule
\end{tabular}
\end{table}

Table~\ref{tab:ablation} examines which components drive the ALFWorld gains
under the 500-token memory budget.

\paragraph{\textbf{Decision utility is the main source of improvement}.}
Replacing utility-based ranking with BM25 reduces SR@50 from $84.33\%$ to
$65.67\%$.
A failure-only heuristic reaches $70.15\%$, indicating that simply
prioritizing failure-related memories does not explain the improvement.
These results highlight the importance of estimating the decision value of
individual memories.

\paragraph{\textbf{Relevance is more useful for recall than arbitration}.}
Directly combining relevance and predicted utility
($\lambda=0.5$) achieves $78.36\%$, $5.97$ points below utility-only
ranking.
This supports our design choice of using relevance for candidate recall while
relying on predicted utility for final memory arbitration.

\paragraph{\textbf{The gain is not a failure shortcut}.}
Removing explicit failure/state indicators further improves SR@50 from
$84.33\%$ to \textbf{$86.57\%$}, which is used as the primary MemCredit
configuration.
Together with the weaker Failure-Aware baseline, this suggests that MemCredit
captures broader state--memory interactions rather than relying on explicit
failure cues.

Overall, the ablation identifies utility-aware arbitration---the
context-conditional credit-per-token selection of
Sec.~\ref{sec:arbitration}---as the main source
of improvement, while direct relevance interpolation and failure-specific
heuristics provide weaker alternatives.
Additional budget analyses are reported in the Appendix.

\section{Conclusion}
We introduced MemCredit, a decision-utility-based memory framework for
long-horizon agents.
Rather than treating semantic relevance as the primary criterion for memory
selection, MemCredit estimates the context-conditional marginal
contribution of individual memories through counterfactual credit
assignment, while keeping the action policy frozen and avoiding
additional environment interaction at deployment.
Experiments on ALFWorld and WebShop show that this principle improves
memory arbitration under limited context budgets.
Our analysis further reveals that decision utility is sparse, strongly
state dependent, and only weakly captured by semantic relevance.
These findings suggest that future agent memory systems should optimize not
only what information is retrieved, but which memories are most consequential
for the decisions that follow.

\bibliographystyle{ACM-Reference-Format}
\balance
\bibliography{sample-base}


\clearpage

\appendix

\section{Additional Experimental Details}
\label{app:experiments}

\subsection{Implementation and Evaluation Details}
\label{app:impl}

\textbf{Hardware and inference.}
All same-model ALFWorld evaluations use Qwen3.6-27B-FP8 served with
vLLM~\cite{kwon2023pagedattention}
on A800-80GB GPUs, with a 4096-token context window, temperature 0, and
seed 42.
All compared methods use the same action protocol, parser, environment,
and admissible-command hints.
The current observation is included in the base prompt and excluded from
the historical-memory budget.

\textbf{Pipeline validation.}
The sharded evaluation pipeline was validated by rerunning BM25@500, which
reproduces the frozen result ($65.67\%$ SR@50) up to the execution-order
variance documented below: across repeated executions, task-level outcomes
agree on at least $94\%$ of the 134 tasks, and single-run aggregate SR
varies by up to $\pm3$~pp.
The agent uses no Python-side randomness; the residual flips trace to
GPU-side logit nondeterminism on near-ties under temperature-0 decoding.
Memory budgets of 500 and 750 tokens apply only to historical memory.

\textbf{Utility label generation.}
Intervention labels follow Eq.~\ref{eq:one_step_utility}: the observed
one-step credit is the difference between the two branches' environment
rewards, $r^{+i}-r^{-i}$, where each branch is executed by the same
frozen policy from the same reconstructed state and differs only in the
rendered inclusion of the target memory.
For ALFWorld, the step reward $r$ is instantiated with an
environment-grounded outcome score taking values in $\{+1,0,-1\}$:
$+1$ if the episode is won at this step; $-1$ if the chosen action is not
in the admissible set or repeats a previously failed action; and $0$ for
a valid but neutral action.
Reference contexts are sampled at occupancy levels
$\alpha\in\{0,\tfrac14,\tfrac12,\tfrac34,1\}$ under the budget, so the
factual branch $A$ and its reference $S=A\setminus\{m_i\}$ cover isolated,
partial, and near-budget contexts.
Each pair is executed once (no repeated sampling), and samples whose
reconstructed state differs from the recorded one are discarded by the
reconstruction check; all 1,041 retained samples pass it.
Primary causal analyses use the single-removal reference context
$S=C_t\setminus\{m_i\}$ for the main statistics.

\textbf{WebShop catalog audit.}
The 10K-product catalog is a uniform subsample of the standard WebShop
catalog.
For all 200 evaluated instructions, the target product and its required
options are present in the catalog, and the target is retrievable within
the top-10 search results under the BM25 index used by the environment
(Recall@10/20/50 = 100\%), so the evaluation is not confounded by missing
catalog entries.
The reward semantics are identical to standard WebShop; because the
catalog is a 10K subset of the full catalog, scores are not directly
comparable to full-catalog literature values.

\subsection{Additional Overall Performance Results}
\label{app:overall}

% Appendix: paired task-level ALFWorld differences vs BM25 (budget 500).
% 131 tasks have valid paired records (three tasks in the sharded utility
% runs were replayed under a different first-step label and are excluded).
\begin{table}[t]
\centering
\caption{
ALFWorld budget-500 comparison relative to BM25 (65.67\% SR@50).
All 134 tasks have valid paired records.
Recovery denotes the post-failure recovery rate and Failed Rep.\ the
failed-action repetition rate; both are taken from the main comparison.
}
\label{tab:paired_appendix}
\small
\begin{tabular}{lccc}
\toprule
Method & $\Delta$SR (pp) & Recovery (\%) & Failed Rep.\ (\%) \\
\midrule
Flat Recency & $-8.21$ & 40.8 & 43.2 \\
Dense Retrieval & $+5.97$ & 53.7 & 32.1 \\
Failure-Aware Retrieval & $+4.48$ & 59.6 & 27.6 \\
State-Aware Retrieval & $+10.45$ & 63.4 & 24.5 \\
MemArbiter & $+17.17$ & 68.2 & 22.4 \\
MemCredit (hybrid) & $+12.69$ & 69.7 & 20.6 \\
MemCredit (utility-only) & $+18.66$ & 72.8 & 18.5 \\
\textbf{MemCredit} & $\mathbf{+20.90}$ & \textbf{75.9} & \textbf{16.4} \\
\bottomrule
\end{tabular}
\end{table}

All 134 ALFWorld tasks have valid paired records in the corrected
pipeline; Table~\ref{tab:paired_appendix} reports each method's budget-500
difference relative to BM25 together with the post-failure behavioral
metrics from the main comparison.
Behavioral metrics reinforce the task-success pattern: methods with larger
SR gains also recover more often after failures and repeat failed actions
less frequently.

\subsection{Additional Causal Analysis}
\label{app:causal}

% Appendix: full causal statistics (combined context-conditional set,
% n=1041, 45 tasks, 100% reconstruction; task-clustered bootstrap 95% CIs).
\begin{table}[t]
\centering
\caption{Decision-utility causal statistics (combined set).
``Decision utility'' is the credit of including one memory item relative
to the reference context; positive means inclusion improves the decision,
and the reported statistics use the single-removal reference context
$S=C_t\setminus\{m_i\}$.}
\label{tab:causal_full_appendix}
\small
\begin{tabular}{lc}
\toprule
Quantity & Estimate [95\% CI] \\
\midrule
Mean memory credit & $+0.054$ [$+0.030$, $+0.080$] \\
Spearman (relevance, utility), combined & $-0.011$ [$-0.073$, $+0.077$] \\
Spearman, dense-only (11 tasks) & $-0.191$ \\
Spearman, broad-only (44 tasks) & $+0.135$ [$+0.062$, $+0.209$] \\
Q1--Q5 relevance-quintile utility gap & $-0.009$ [$-0.072$, $+0.069$] \\
Utility zero-share & 0.869 \\
Failure-context utility & $+0.176$ [$+0.077$, $+0.257$] \\
Non-failure utility & $+0.029$ [$+0.005$, $+0.057$] \\
Harmed items in semantic top-1 / top-3 / top-5 & 0.000 / 0.000 / 0.000 \\
\bottomrule
\end{tabular}
\end{table}

\begin{table}[t]
\centering
\caption{Mean memory credit by state type (combined set).}
\label{tab:states_appendix}
\small
\begin{tabular}{lcc}
\toprule
State type & Mean credit [95\% CI] & Zero share \\
\midrule
failure\_context & $+0.176$ [$+0.077$, $+0.257$] & 0.722 \\
normal\_progress & $+0.060$ [$+0.007$, $+0.120$] & 0.863 \\
repeat\_risk & $+0.035$ [$-0.019$, $+0.080$] & 0.923 \\
constraint\_sensitive & $+0.033$ [$-0.010$, $+0.079$] & 0.900 \\
reference\_sensitive & $+0.023$ [$-0.023$, $+0.076$] & 0.853 \\
subgoal\_transition & $-0.005$ [$-0.036$, $+0.031$] & 0.962 \\
\bottomrule
\end{tabular}
\end{table}

The combined causal set covers a dense task subset and a broader
task-diverse subset, as summarized in
Table~\ref{tab:causal_full_appendix};
Table~\ref{tab:states_appendix} additionally reports mean memory credit by
state type.
Primary uncertainty estimates use task-clustered bootstrap confidence
intervals.
We report dense-only and broad-only statistics separately to assess
sensitivity to task coverage.
Additional ranking-based diagnostics are treated as exploratory because
positive-credit items are sparse and candidate sets can be small.

\subsection{Utility Prediction Details}
\label{app:prediction}

\textbf{Evaluation protocol.}
Utility predictors are evaluated using 5-fold GroupKFold split by task ID,
preventing intervention samples from the same task from appearing in both
training and evaluation folds; main results are reported in
Table~\ref{tab:utility_prediction}.

\textbf{Features.}
Relevance-only features contain semantic relevance and recency age;
failure-only features contain previous-action failure, failure-related
indicators, and state-type features;
content features contain inference-time memory and state descriptors.
The full estimator adds the context descriptors $x_{\mathrm{ctx}}$
(token-level overlap with the selected context, candidate-entity coverage,
per-type context counts, and budget occupancy), which expose whether a
candidate contributes information already represented in the selected
context.
A context-blind control omits these descriptors.
No counterfactual outcome or intervention-result fields are available to the
predictor.

\textbf{Estimator.}
The primary ExtraTrees~\cite{geurts2006extratrees} model uses 300 trees,
maximum depth 8, and minimum leaf size 8.
Ridge regression with $\alpha=1$ is used as a simpler control.

\textbf{NDCG diagnostic.}
NDCG@5 is omitted from the main text because $86.9\%$ of utility labels are
exactly zero, producing extensive ranking ties.
The shuffled-label control obtains NDCG@5 $0.301$, compared with $0.131$ for
the full estimator, indicating that this metric is not informative under the
current label distribution.
We therefore rely primarily on Spearman correlation, AUPRC, and AUROC.

\subsection{Additional Ablations}
\label{app:ablations}

The hybrid ablation variant uses a fixed interpolation coefficient
$\lambda=0.5$, selected before final evaluation.
The Failure-Aware baseline adds a fixed bonus to failure-related memories
while preserving the same retrieval and memory budget.
No post-hoc tuning is performed on the final 134-task evaluation set.

A larger 750-token budget raises utility-only ALFWorld SR from $84.33\%$
to $88.81\%$, and the primary MemCredit configuration from $86.57\%$ to
$90.30\%$.
Additional failure-density analyses stratify tasks using bins fixed from
baseline trajectories before comparing methods.

\subsection{Performance across Failure Density}
\label{app:failure_density}

We further examine whether the benefit of MemCredit is restricted to
failure-intensive trajectories.
Using the frozen BM25 trajectories, we compute the failure density of each
ALFWorld task and partition the 134 tasks into low-, medium-, and
high-failure-density terciles.
The resulting partition is fixed and applied to both BM25 and MemCredit,
avoiding method-dependent grouping.

\begin{figure}[t]
    \centering
    \includegraphics[
        width=\columnwidth
    ]{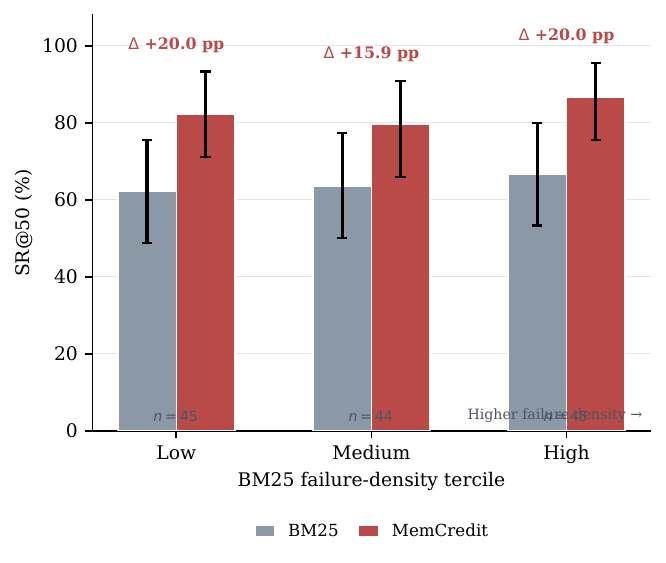}
    \caption{
    \textbf{ALFWorld performance across baseline failure density.}
    Tasks are partitioned into terciles using failure density measured from
    frozen BM25 trajectories, and the same partition is used for both methods.
    MemCredit consistently improves over BM25 across all three groups.
    Error bars denote task-bootstrap 95\% confidence intervals; annotations
    report paired SR@50 differences and their 95\% confidence intervals.
    }
    \label{fig:failure_density}
\end{figure}

As shown in Figure~\ref{fig:failure_density}, MemCredit consistently
outperforms BM25 across all three groups.
The SR@50 improvements are $+20.0$, $+15.9$, and $+20.0$ percentage points
for the low-, medium-, and high-failure groups, respectively.
The paired task-bootstrap 95\% confidence intervals remain above zero in all
three terciles: $[+6.7,+33.3]$, $[+4.5,+29.5]$, and $[+8.9,+33.3]$.

This result complements the intervention analysis in
Sec.~\ref{sec:utility}.
Although individual memories exhibit substantially larger marginal utility
in failure contexts, the end-to-end advantage of utility-aware arbitration
is not confined to failure-heavy tasks.
Failure context therefore represents an important source of memory utility,
but not the only setting in which decision-aware memory selection is useful.

\end{document}